%% file: root.tex
\documentclass[letterpaper, 10 pt, conference]{ieeeconf}  

\IEEEoverridecommandlockouts                              
\usepackage{graphicx} 
\usepackage{url}
\usepackage{caption}
\usepackage{subcaption}
\usepackage{amsmath}
\usepackage{amsfonts}
\usepackage{mathrsfs}  
\usepackage{enumerate}
\usepackage{algpseudocode}
\usepackage{cite}
\usepackage{float}
\usepackage[ruled,vlined]{algorithm2e}                                                       

\usepackage{amsmath}

\title{\LARGE \bf
Deep Predictive Policy Training using Reinforcement Learning
}

\author{Ali Ghadirzadeh,  Atsuto Maki, Danica Kragic and M{\r a}rten Bj\"orkman
\thanks{Authors are with the Robotics, Perception and Learning Lab (RPL), CSC, KTH Royal Institute of Technology, Stockholm, Sweden.
        {\tt\small algh|atsuto|dani|celle@kth.se}}%
}
\begin{document}

\maketitle
\thispagestyle{empty}
\pagestyle{empty}

\input{texts/abstract}
\input{texts/introduction.tex}
\input{texts/related_work.tex}
\input{texts/background.tex}
\input{texts/background_autoencoder.tex}
\input{texts/background_spatial_autoencoder.tex}
\input{texts/background_variational_autoencoder.tex}
\input{texts/method.tex}
\input{texts/method_perception.tex}
\input{texts/method_behavior.tex}
\input{texts/method_action_policy.tex}
\input{texts/experiments.tex}
\input{texts/conclusions.tex}

\addtolength{\textheight}{-12cm}   



 
\vspace{-1.2 mm}
\section*{ACKNOWLEDGMENT}
This work was supported by the EU through the project socSMCs (H2020-FETPROACT-2014) and the Swedish Research Council. 
\vspace{-1.2 mm}

\bibliographystyle{IEEEtran}
\bibliography{root}

\end{document}

%% file: texts/abstract.tex
\begin{abstract}
Skilled robot task learning is best implemented by predictive action policies due to the inherent latency of sensorimotor processes. However, training such predictive policies is challenging as it involves finding a trajectory of motor activations for the full duration of the action.
We propose a data-efficient deep predictive policy training (DPPT) framework with a deep neural network policy architecture which maps an image observation to a sequence of motor activations.
The architecture consists of three sub-networks referred to as the perception, policy and behavior super-layers.
The perception and behavior super-layers force an abstraction of visual and motor data trained with synthetic and simulated training samples, respectively. 
The policy super-layer is a small sub-network with fewer parameters that maps data in-between the abstracted manifolds. It is trained for each task using methods for policy search reinforcement learning.
We demonstrate the suitability of the proposed architecture and learning framework by training predictive policies for skilled object grasping and ball throwing on a PR2 robot. The effectiveness of the method is illustrated by the fact that these tasks are trained using only about 180 real robot attempts with qualitative terminal rewards.
\vspace{-0.4 cm}
\end{abstract}

%% file: texts/introduction.tex
\section{Introduction}
\label{sec:introduction}
We humans are \textit{skilled} in a majority of our basic physical activities such as opening a door or grasping an object, and also demonstrate impressive motor learning abilities to acquire new skills e.g., learning to play a new sport.
On the other hand, most robotic systems demonstrate behaviors far from being considered \textit{skilled}, especially in unstructured environments. 
The gap between humans and robots in motor skill learning may be explained not only by the highly versatile sensing and actuating capabilities of humans, but also by the way the sensorimotor process is intertwined.

Studies in motor learning of biological systems reveal that skilled action performance is likely to be the result of predictive types of controllers, i.e., an uninterrupted motor activation executed for a given observation snapshot \cite{wolpert2011principles}. 
This differs from the reactive types of controllers which produces motor activation in response to every sensory input and inherently slows down the process by at least the sensor delays. 
Task execution for basic physical activities may thus be unnecessarily slow on robotic systems operating with reactive controllers.
This can be improved by adapting predictive action policies to the robotic systems with an efficient motor learning process.

\begin{figure}[h]    
  \centering
    \includegraphics[width=0.47\textwidth]{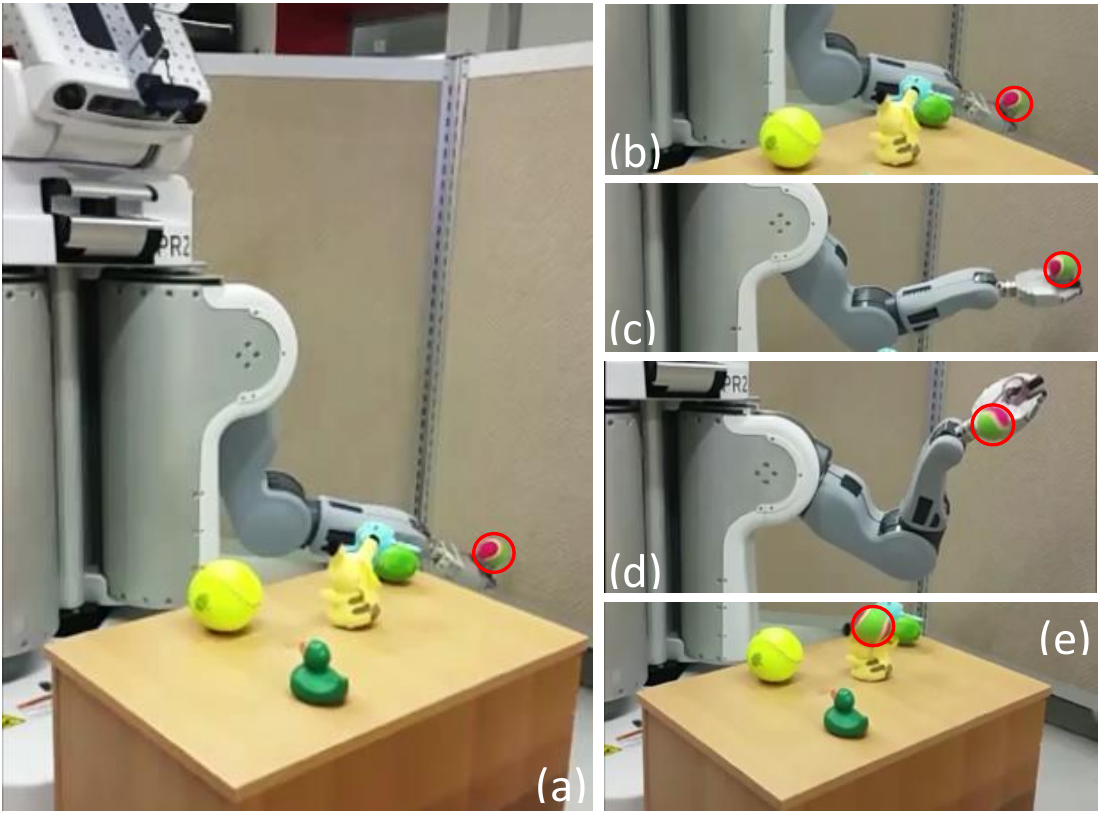}    
    \caption{The experimental setup for the ball throwing task (a), and four snapshots demonstrating a successful throw towards a target object (b-e). The ball is highlighted with red circles to improve the visibility.} 
    \label{fig:pr2throw} 
    \vspace{-0.6
    cm}
\end{figure}

\begin{figure}[h]
  \vspace{0.2 cm}
  \centering
    \includegraphics[width=0.48\textwidth]{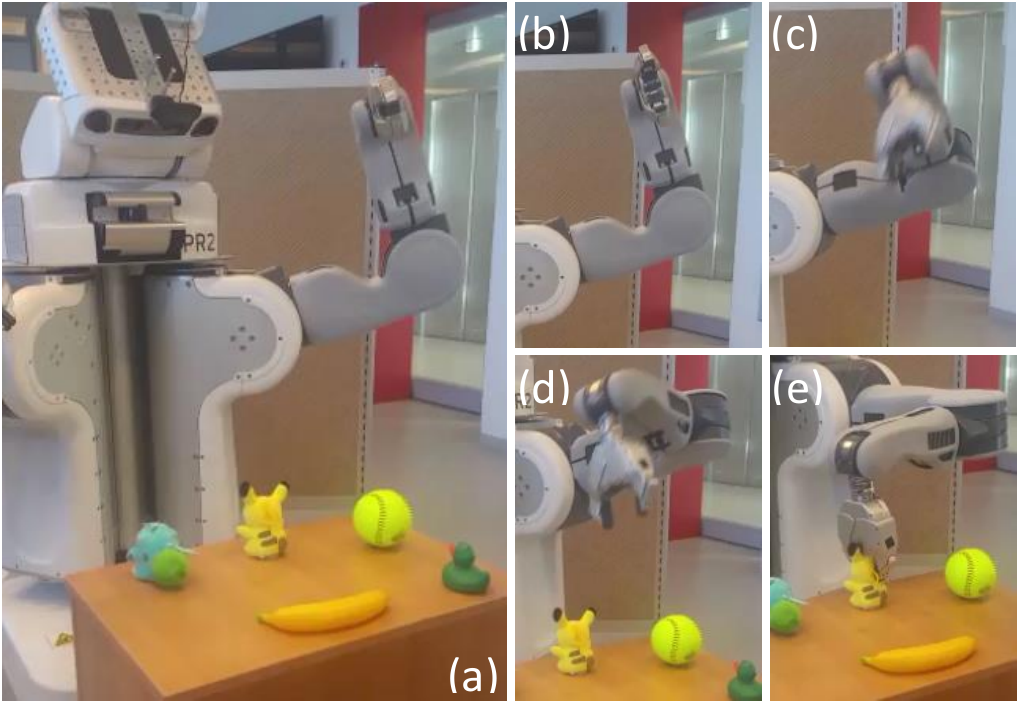}    
    \caption{The experimental setup for the grasp task (a), and four subsequent snapshots demonstrating a successful predictive object grasping (b-e).} 
    \label{fig:pr2grasp}	
    \vspace{-0.5 cm}
\end{figure}

In this work, we present a framework based on reinforcement (see \cite{kober2013reinforcement,deisenroth2013survey} for surveys) and use-dependent learning \cite{wolpert2011principles} to train a deep neural network policy, mapping uncalibrated image observations to long trajectories of low-level motor commands for a robotic manipulator. 
The main contributions are: 
1) a robot learning framework to acquire skilled behaviors, 
2) a neural network architecture for data-efficient deep policy learning consisting of three super-layers (see Fig.~\ref{fig:network}): two for abstraction of visual data and motor trajectories  and one for mapping data in-between the abstracted manifolds, 
and 3) a mechanism that enables learning a behavior in simulation and exploiting it on a real robot. 
The framework is applied on a PR2 robot to learn two skilled behaviors, ball throwing and object grasping, as shown in Fig.~\ref{fig:pr2throw} and Fig.~\ref{fig:pr2grasp}. 

We demonstrate experimentally that the network architecture enables training of complex behaviors with qualitative terminal rewards,
rewards provided at the end of trajectories, evaluating whole sequences with qualitative measures such as "good" or "excellent".
This reward type is less informative, but it is often more realistic in practice, especially for training predictive policies, due to the difficulties of evaluating every single time-step action and the latency of the reward system. 
Furthermore, providing qualitative rewards require less engineering efforts and can be done by a non-expert operator during a training phase.

This paper is organized as follows: In the rest of this section, we provide a short background, as well as review earlier work related to the presented. Sec.~\ref{sec:representation_learning} introduces two structures to abstract motor and image data. The learning framework as well as the architecture is presented in Sec.~\ref{sec:method}. Experimental results are provided in Sec.~\ref{sec:experiments}, testing one simulated task and two real robotic tasks. Finally, Sec.~\ref{sec:conclusions} will conclude our discussions and suggest future work. 
\subsection{Background}
In general, a deep action policy consists of a mechanism to extract informative states from raw sensor data, and another mechanism to generate motor activations (or sequences of activations, in the case of predictive policies) for each state. 

Given its richness as a sensory modality, vision plays an important role in providing information about the environment and the state of the system. However, visual data may contain more information than can be processed efficiently by an agent that learns a motor behavior.
Therefore, due to processing limitations, the agent needs to filter out data redundant to the motor learning task. 
This phenomenon is known as inattentional blindness in psychological studies \cite{most2005you}. 
Traditionally in robotics, filtering unnecessary information to extract a task-relevant state representation is done using hand-crafted features (e.g., \cite{ghadirzadeh2016sensorimotor}). 
However, in recent years a considerable amount of research is devoted to learning state representations from raw observation data (e.g.,~\cite{finn2015deep,lange2012autonomous,jonschkowski2014state,watter2015embed,wahlstrom2015learning,van2016stable}). This  improves autonomy of motor task learning. Furthermore, simultaneously training perception and action policy may result in an overall better policy, compared to hand-crafted perception models \cite{levine2016end}. 



The other mechanism is required to learn to control motor activations to perform a task. 
Studies in neuroscience \cite{wolpert2011principles} distinguish between three types of motor learning processes in biological systems: 1) error-based learning, 2) reinforcement learning, and 3) use-dependent learning. Error-based learning optimizes each motor activation in each time-step by updating it in the opposite direction of the error gradient (see e.g., \cite{ghadirzadeh2015sensorimotor,ghadirzadeh2016self}). The error is generally found internally as the difference between the perceived sensory outcome to the desired or predicted one. 
Reinforcement learning (RL) improves the action selection policy by reinforcing those actions which are likely to yield higher rewards. 
The reward is given by the environment, including the robot itself and possibly an instructor. 
Finally, use-dependent learning is a learning process based on movement repetitions without a target. It facilitates policy training by modeling correlations in motor activations i.e., encoding kinematic details of different motions. This enables the other two processes to start training at a higher level abstraction of motor activations.

%% file: texts/related_work.tex
\subsection{Related work}
\label{sec:related_work}
\begin{figure*}[t!]
	\vspace{0.3cm}
    \centering
    \includegraphics[width=1\textwidth]{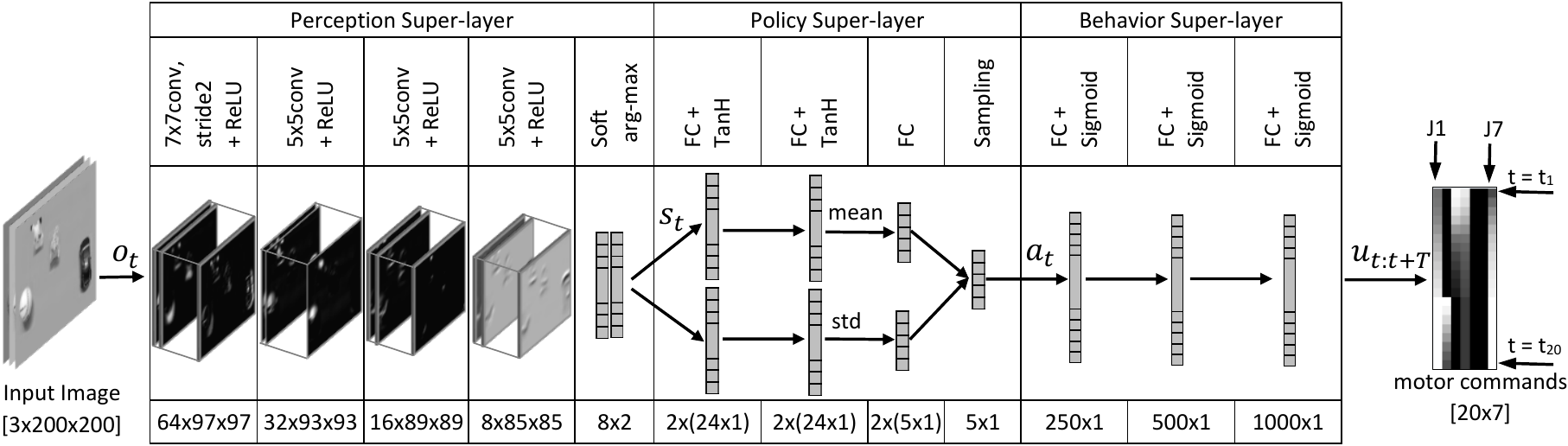}
    \caption{The deep predictive policy architecture consisting of the perception, policy and behavior super-layers. A mean-centered RGB image is given as the network input. The perception super-layer abstracts the image data with a number of spatial positions corresponding to the task-related objects. The policy super-layer stochastically maps the abstracted state into a point in the action manifold. Finally, the behavior super-layer generates a long trajectory of motor commands for the given sampled action that is applied to the robot for $T$ consecutive time-steps.}
    \label{fig:network}	
    \vspace{-0.7 cm}
\end{figure*}  
In this section, we review related studies regarding 1) methods for extracting  task-relevant state representation from raw sensory observations, and 2) methods for predictive action policy learning. 
\input{texts/related_work_perception.tex}
\input{texts/related_work_policy.tex}

%% file: texts/related_work_perception.tex
\subsubsection{State representation learning}
Different methods have earlier been applied in robotics to infer a compact state representation from high-dimensional observations, in particular camera images, applicable to motor control learning tasks. 
The study conducted by Jodogne and Piater \cite{jodogne2007closed} is among earlier studies on task-specific state representation learning. Their method finds features that distinguish a finite set of classes based on which the agent makes action decisions. However, their method may not be directly applicable to many real robotic problems, as discretizing of sensory observations into a finite number of classes is not always possible. 
In a more recent study, Jonschkowski and Brock \cite{jonschkowski2014state} proposed learning a task-relevant state representation based on prior physical knowledge. 
Although being a promising concept, their approach has so far only been validated on a toy example and may not be data-efficient enough to be applied on real robotic problems. 

Training a low-dimensional state representation in an autoencoder structure (refer to Sec.~\ref{sec:representation_learning}) has gained popularity in recent years \cite{finn2015deep,lange2012autonomous,jonschkowski2014state,watter2015embed,wahlstrom2015learning,van2016stable}.
Lange et al.~\cite{lange2012autonomous} trained a deep convolutional autoencoder to learn a state space representation to autonomously control a toy race car. 
They devised a transformation based on expert knowledge to shape the state manifold for the control task. 
In similar work, Wahlstr{\"o}m et al.~\cite{wahlstrom2015learning} used an autoencoder structure to find a state manifold based on which a multi-time-step prediction of camera images is feasible. 
For this purpose, they refined the cost function to include the reconstruction error for the current and the next time-step.  
Watter et al.~\cite{watter2015embed} proposed a similar cost function to train a variational autoencoder (refer to Sec.~\ref{sec:variational_autoencoder}) with an extra constraint that state dynamics are to be linearizable w.r.t. all control signals. The same approach is also used by van Hoof et al.~\cite{van2016stable} to extract task-specific state representations for visual and tactile data. 
However, methods to augment specific types of dynamics simultaneously with state representation learning may be impractical in certain tasks, such as hitting a table-tennis ball, where semi-random motor actions sparsely result in sensible task-relevant outcomes during the initial training phase.

Alternatively, certain properties of the states' dynamics can be ensured by limiting the autoencoder to a specific class of features e.g., spatial image features. 
Based on this idea, Finn et al.~\cite{finn2015deep} proposed a deep spatial autoencoder for visuomotor tasks. 
They exploited the spatial softmax layer, introduced in their earlier work \cite{levine2016end}, to convert the activation of the last layer of the convolutional filters into spatial image positions. 
This topology has been  applied in a number of real robotic visuomotor learning tasks \cite{levine2016end,chebotar2016path,agrawal2016learning,yahya2016collective}.

Beyond methods of finding a low-dimensional state representations suitable for action policy learning, there are other methods that can train perception and policy end-to-end, without a clear-cut boundary. 
The most well-known examples are the Deep Q-network (DQN) \cite{mnih2015human} and deep deterministic policy gradient (DDPG) \cite{lillicrap2015continuous} methods for discrete and continuous action spaces, respectively. 
In general, these methods are hardly applicable to robotics problems, since they require a large amount of agent-environment interaction data which may not be affordable with real robotic setups. 

Another approach to train perception-policy end-to-end is the guided policy search (GPS) introduced by Levine et al.~\cite{levine2016end}.
Guided policy search is a framework which converts policy search reinforcement learning into a supervised learning paradigm with the supervised data coming from a secondary trajectory optimizer such as iterative linear-Gaussian regulator (iLQG) \cite{levine2014learning}. 
GPS may be used in training a deep predictive policy; however it would require  a secondary trajectory optimizer method with an engineered initial state representation. 

In this work, we exploit spatial autoencoders introduced in \cite{finn2015deep} to learn a low-dimensional state representation. 
This is a reasonable choice, since task-relevant states can be learned in an autoencoder without requiring the robot to actively manipulate the environment. 
We improve the spatial autoencoder stability w.r.t. visual distractors by further training the convolutional layers to filter out these distractors.

%% file: texts/related_work_policy.tex
\subsubsection{Predictive action policy learning}
Predictive actions can be found using optimal control theory methods. These methods optimize a known cost function given a set of differential equations and constraints. 
The differential equations represent the dynamic model of the system, and the constraints determines limitations on the control signals and states. 
The methods in optimal control theory most resembling our approach are trajectory optimization \cite{betts2010practical} and model predictive control (MPC) \cite{camacho2013model}. 

Trajectory optimization methods find open-loop control trajectories by optimizing a given cost function for a given dynamic model. 
Similarly, MPC optimizes a cost function over a finite horizon. However, unlike trajectory optimization, MPC applies only the first action of the trajectory and repeats this procedure in every time-step. 
Our approach differs from optimal control theory  in that ours is data-driven and is not dependent on a known dynamic model. 

More recently, MPC is accompanied by general function approximators, such as Gaussian processes \cite{kocijan2004gaussian,ghadirzadeh2016self} or artificial neural networks \cite{wahlstrom2015learning} to learn the dynamic model directly from interaction data to alleviate the need for an a-priori known dynamic model. 
However, learning a dynamic model from data may not always be feasible, especially in high-dimensional sensorimotor spaces, and the error, the difference between the true dynamics and the predicted ones, can be accumulated over the prediction horizon resulting in an incorrect control trajectory. Furthermore, MPC still reactively responds to each state perceived by the system and requires an expensive optimization in each time-step. 

Another related approach is a biologically inspired method \cite{wolpert2011principles} to split a complex behavior into a set of basic motor primitives. These approaches, known as dynamic movement primitives (DMP), have found applications in robot learning by demonstration \cite{ijspeert2003learning} and also reinforcement learning \cite{kober2009policy,deisenroth2013survey}.
They have been successfully applied to a number of robotic skilled behavior learning tasks, such as robot tennis swings \cite{ijspeert2003learning}, ball-in-a-cup and underactuated swing-up \cite{kober2009policy}, dart throwing and table tennis \cite{kober2010reinforcement}. 

Our method relates to DMP in that both methods find a representation of motor trajectories to train high-level behaviors. 
However, in our case, each point in a 5D action-manifold, passed through a fully-connected neural network, characterizes a motion trajectory for all the robot joints,
while in the DMP case, the motion of each joint is characterized by a set of differential equations with a number of trainable parameters. 
Our method compared to DMP makes a more abstract representation while being computationally less expensive. 

%% file: texts/background_autoencoder.tex
\section{Representation learning}
\label{sec:representation_learning}
In order to train the deep policy in a data-efficient manner, the mapping from high-dimensional observations $o_t$ to motor trajectories $u_{t:t+T}$ is done in a low-dimensional space by abstracting observation and motor data with $s_t = f_p(o_t)$ and $u_{t:t+T} = g_b(a_t)$, where $s_t$ and $a_t$ are the abstracted state and action manifolds and the two functions, $f_p(.)$ and $g_b(.)$, represent the perception and behavior super-layers as illustrated in Fig.~\ref{fig:network}.
These super-layers are trained  by two different structures, based on spatial \cite{finn2015deep} and variational \cite{kingma2013auto} autoencoders, introduced later in this section.

An autoencoder is an hourglass-shaped artificial neural network which learns a low-dimensional representation of the training data at its bottleneck. 
It learns such representations by first mapping the input data into the low-dimensional manifold (encoder) and then reconstructing the input from the low-dimensional data at the network output (decoder).  
We train two autoencoder structures to find an abstract data representation for two purposes:
1) to efficiently extract task-relevant states from raw camera images through the perception super-layer, and
2) to infer motor signal data distributions for a skilled behavior to be able to produce samples as close as possible to the learned distribution. 
In following subsections, we introduce the two autoencoder structures used on our framework. 

%% file: texts/background_spatial_autoencoder.tex
\subsubsection{Convolutional spatial autoencoders \cite{finn2015deep}}
\label{sec:spatial_autoencoder}
This autoencoder encodes the input image with the 2D positions of a number of points belonging to the task-related objects of the scene. The input image is reconstructed based on this encoding i.e., based on the knowledge of where the relevant objects are located in the image. The encoding inherently preserves spatial distances in the input image and is therefore suitable for robotic manipulation tasks \cite{finn2015deep,levine2016end,chebotar2016path,yahya2016collective}.  

The encoder of a convolutional spatial autoencoder consists of concatenations of several convolutional layers followed by a spatial soft arg-max.
For a given input, each convolutional filter generates a 2D response map. The encoded features correspond to the response map of the last convolutional layer. These maps are first normalized and transformed into probability density functions based on the spatial softmax layer, as $s_{i,j}^c = \exp{(\frac{\sigma_{i,j}^c}{\alpha})} / \sum_{i',j'}{\exp{(\frac{\sigma_{i',j'}^c}{\alpha})}}$, where $\sigma_{i,j}^c$ is the $(i,j)$ element of the $c_{th}$ response map and $\alpha$ is a trainable temperature parameter. The encoded feature point is found as $(p^c_x, p^c_y) = (\sum_{x',y'}{s^c_{x',y'}x}, \sum_{x',y'}{s^c_{x',y'}y})$ for each filter $c$. Therefore, the size of the encoded feature space is twice the number of the filters in the last convolutional layer. The decoder part is a fully-connected neural network which reconstructs a down-sampled gray-scale version of the input image. Please refer to the original work \cite{finn2015deep} for a more detailed description. 

%% file: texts/background_variational_autoencoder.tex
\subsubsection{Variational autoencoders \cite{kingma2013auto}}
\label{sec:variational_autoencoder}
A generative behavior model is trained to represent long motor trajectories with a low-dimensional action manifold. In this way, a motor task can be learned by searching for a policy in the action manifold instead of the high-dimensional motor trajectory space, making the search considerably more efficient.
However, in order to benefit from policy search in the low-dimensional action manifold, we need to ensure the action data is encoded with a proper distribution. 
Variational autoencoders reproduce training data 
by sampling a latent variable (in our case the action $a$) from a prior distribution $p(a)$,  typically an isotropic Gaussian $p(a) = \mathcal{N}(a| 0, I)$ \cite{doersch2016tutorial,kingma2013auto}.

The key idea is to learn an encoder $f_b(a | u)$ and decoder $g_b(a)$, where $f_b(a | u)$ for an input $u$, gives a distribution over $a$ with values that are likely to regenerate $u$ when applied to $g_b(a)$. 
The encoder outputs are typically assumed to be normal distributed, $f_b(a|u) = \mathcal{N}(a|\mu(u), \Sigma(u) )$. In essence the encoder consists of two parts; a mean network $\mu(u)$ and a variance network $\Sigma(u)$.
To make $a$ distributed according to the prior distribution $p(a)$, the Kullback-Leibler divergence (KL-divergence) between $f_b(a|u)$ and $p(a)$ is defined as an extra loss function, 
\begin{equation}
\mathcal{L}_{d} = D_{KL}(\mathcal{N}(a|\mu(u), \Sigma(u) )||\mathcal{N}(a| 0, I)),
\end{equation}
where it is assumed that the prior distribution is an isotropic Gaussian and $D_{KL}$ represents the KL-divergence distance. 

%% file: texts/method.tex
\section{Training a deep predictive policy}
\label{sec:method}
We describe our method to train the deep predictive policy architecture consisting of the perception, policy and behavior super-layers shown in Fig.~\ref{fig:network}.
The input image $o_t$ is processed by the perception layers to output a number of spatial image points representing the image position of the task-relevant objects to form the state vector $s_t$. 
The policy super-layer processes the state $s_t$ and produces a normal distribution from which the action $a_t$ is sampled. 
The sampled action $a_t$ is mapped to a predictive trajectory of $T$ time-step motor outputs $u_{t:t+T}$ by the behavior super-layer. 
 
The perception and behavior layers are trained individually in two different autoencoder structures. The encoder network of the perception autoencoder $f_p(.)$ and the decoder network of the behavior autoencoder $g_b(.)$ are then used as the perception and behavior super-layers of the deep predictive policy, respectively. 
Finally, after the perception and behavior layers are set, the policy layers are trained based on RL policy search. 
It is important to emphasize that, given the abstraction of both perception and action, the input-output of the policy super-layer is typically low-dimensional and it is quite affordable to train the few parameters of this small sub-network with standard RL policy search.  
In the rest of this section, we provide details on how to train each of the super-layers to learn a skilled behavior. 

%% file: texts/method_perception.tex
\subsection{Perception super-layer}
\label{sec:method_perception}
The perception model is trained by the spatial autoencoder \cite{finn2015deep} introduced  in Sec.~\ref{sec:spatial_autoencoder}.
As shown in Fig.~\ref{fig:network}, the perception model processes a $200\times200$ mean-centered RGB image with a concatenation of four convolutional layers followed by a spatial soft arg-max.  
There are $8\times2$ image coordinates corresponding to the 8 filters of the last convolutional layer. 

The decoder network consists of two fully-connected hidden-layers with 500 and 2000 neurons, respectively. The decoder output layer formed by 3600 neurons to reconstruct the gray-scaled version of the input image with $60\times60$ pixels. 
Following the original work, we added an extra cost to penalize the variability of learned features for consecutive training data, when images come in sequence.

\begin{figure}[h]    
  \centering
    \includegraphics[width=0.43\textwidth]{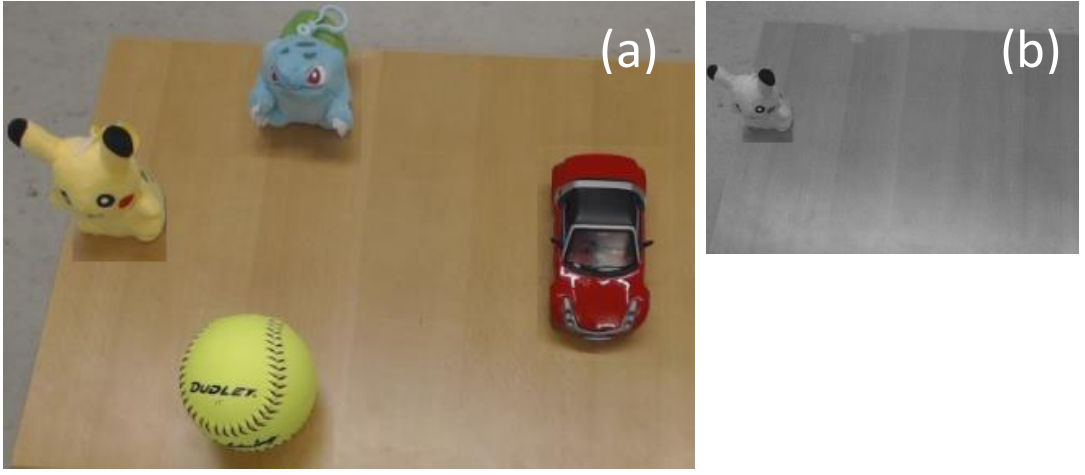}    
    \caption{Generating a synthetic image database to train the perception autoencoder. A sample synthetic image input (a), and the corresponding target image (b).} 
    \label{fig:pikachu}
    \vspace{-0.4 cm}
\end{figure}

The perception layer encodes the task-relevant object position with 8 image points. For the experiments in Sec.~\ref{sec:experiments} we used the Pikachu toy shown in Fig.~\ref{fig:pikachu} as target. A database of 50 images was generated with the Pikachu in different places over the table. A synthetic database containing 5400 images was created by randomly displacing the Pikachu within few pixels and also overlaying a number of distractors at random positions, as shown in Fig.~\ref{fig:pikachu}a. 
The autoencoder is trained to discard the irrelevant distractors by only reconstructing the task-specific object (Fig.~\ref{fig:pikachu}b). 
This is similar to the idea of denoising autoencoders but instead of adding noise to the input image, different visual distractors are superimposed at random image positions. In this way, the convolutional filters are  trained to discard the distractors while giving a maximum activation for the task-specific object.

%% file: texts/method_behavior.tex
\subsection{Behavior super-layer}
\label{sec:method_behavior}
As shown in Fig.~\ref{fig:network}, the behavior super-layer is a generative model which maps a $5D$ point $a_t$ from the action-manifold into a motor trajectory $u_{t:t+T}$ consisting of $T=20$ time-steps for the $7$ joints. 
A valid motion trajectory of a robotic manipulator consists of highly correlated motor outputs to realize a specific behavior. 
The correlation is due to the smoothness of the end-effector motion in task space which is governed by the kinematic structure of the robot and the low dynamics of the motor system.
Therefore it is natural to assume that the motor outputs follows a specific distribution $P_{b}$ which is unknown a-priori. 
It is the purpose of the variational autoencoder described in Sec.~\ref{sec:variational_autoencoder} to capture this distribution given past motor activations.

The encoder of the variational autoencoder transforms input trajectories into a normal distribution in a $5D$ space using a neural structure with three layers of 1000, 500 and 250 hidden units for both the mean and variance networks, $\mu(u)$ and $\Sigma(u)$, that constitute the encoder. The two first layers are shared between the networks. A sample is drawn according to the encoded mean and variance and is mapped by the decoder to a complete trajectory. The decoder, that will represent the behavior super-layer, has three hidden layers similar to the encoder, but in reverse order. 

Trajectory data to train the variational autoencoder may not necessarily come from the real robot. 
In our experiments, we produced the training data in a simulated environment. 
Although the simulated robot does not behave exactly like the real robot, 
it is assumed that the action manifolds in simulation and on the real robot are related by a transformation, a transformation that will be captured by the policy super-layer that is yet to be learned.
This claim agrees with our experimental results presented in Sec.~\ref{sec:experiments}. 

\begin{figure}[h]    
  \centering
    \includegraphics[width=0.485\textwidth]{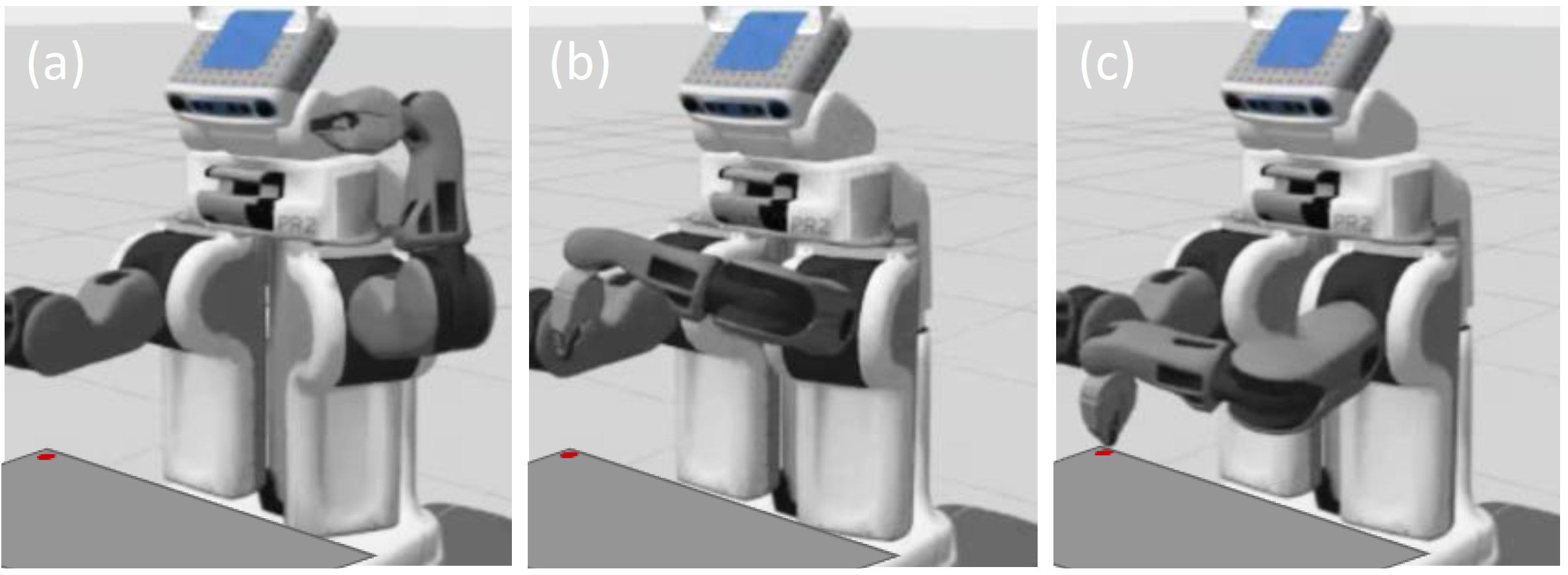}    
    \caption{Three consecutive snapshots showing a successful predictive reaching task in Gazebo simulator (a-c).} 
    \label{fig:simulator}
    \vspace{-0.5cm}
\end{figure}

A blind action policy is required to generate simulated motor trajectories,  based on which the autoencoder is trained.
A blind policy is a controller which randomly generates action trajectories irrespective of state. 
We devised such controllers for each task by demonstrating several motion samples to the robot and deriving a general model which produces similar motions. 
The use of \textit{blind} controllers is also common in other studies e.g., \cite{finn2015deep} to capture initial training data. Here, we only apply the blind controller on the simulated robot. 
We trained two behavior models corresponding to the throwing and grasping behaviors with 10000 motor trajectory samples gathered with the Gazebo simulator shown in Fig.~\ref{fig:simulator}. 

%% file: texts/method_action_policy.tex
\subsection{Policy super-layer}
\label{sec:action_policy}
A predictive action policy learning task can be represented as a MDP problem. The goal is to maximize the expected reward $\mathbb{E}p_\tau[r]$ over trajectories $\tau =\{o_t, u_{t+1}, ..., u_{t+T}\}$, with $o_t$ being the observation at time $t$ and $\{u_{t+1}, ..., u_{t+T}\}$ the motor outputs predictively determined for $T$ time-steps. A trajectory reward $r_{t+T+1}$ is given at the end of each episode as a discrete or a continuous value. 
The predictive policy determines the probability distribution over $T$ time-step motor commands given the observation at time $t$, $\pi(o_t)=p(u_{t+1},..., u_{t+T}| o_t)$. The distribution of trajectories can thus be written as $p_\tau= \pi(o_t)p(o_t)$. 
The goal is to find the policy parameters such that the likelihood of trajectories with higher terminal rewards increases.

As discussed in Sec.~\ref{sec:representation_learning}, instead of training a deep policy $\pi(o_t)$ in the high-dimensional sensorimotor space, we train a low-dimensional policy $\pi'(s_t)$ in the state-action manifold.  Algorithm~\ref{alg:policy_learning} summarizes how to train this low-dimensional policy. 
The perception and behavior super-layers are trained separately with synthetic and simulated data as discussed in Sec.~\ref{sec:method_perception} and Sec.~\ref{sec:method_behavior}, prior to policy training. 
The policy super-layer is initialized such that for any input state, the output is distributed as $\mathcal{N}(0,I)$. 
For each episode, an input image observation is captured and encoded to the state manifold with the perception super-layer. 
Given the input state, the policy $\pi'(s_t)$ generates a distribution over the action manifold from which a sample is drawn. The sampled action is then mapped to the corresponding motor trajectory through the generative behavior super-layer. The motor commands are applied on the robot for $T$ time-steps and the reward is received at the end of each episode. 
The policy is trained with the training data $\{s_i, a_i, r_i\}_{i=1:N_e}$ collected after $N_e$ number of episodes to maximize the expected reward $\mathbb{E}p_\tau[r]$. 


We evaluated RL policy search including vanilla policy gradient (VPG), relative entropy policy search (REPS), cross entropy method (CEM) (see \cite{deisenroth2013survey,kober2013reinforcement} for surveys) and trust region policy optimization (TRPO) \cite{schulman2015trust} for the simulated task (refer to the Sec.~\ref{sec:simulated_tasks}).
Based on the observed convergence rates achieved in the simulations, we chose TRPO to train the policies for the real-robot tasks. 
\begin{algorithm}[h]
\SetKwInOut{Input}{Input}
\SetKwInOut{Output}{Output}
\caption{Training the policy super-layer.}
\Input{Trained perception and behavior super-layers}
\Output{Trained policy super-layer}
Initialize policy $\pi'$ with $\mathcal{N}(0,I)$\;
\For{each iteration}{
  \For{each episode}{
      Input an image observation $o_t$\;
      $s_t \leftarrow f_p(o_t)$\; 		    
      Sample $a_t \sim \pi'(s_t)$\;
      $u_{t:t+T} \leftarrow g_b(a_t)$\;
      Run $u_{t:t+T}$ on the robot\;
      Input episode reward $r_{t+T+1}$\;
      Record the triple $\{s_t, a_t, r_{t+T+1}\}$\;
  }
  Train the policy $\pi'$ with the recorded triples\;
}
\label{alg:policy_learning}
\end{algorithm}

%% file: texts/experiments.tex
\section{Experiments}
\label{sec:experiments}
We devised three different experiments to evaluate the suitability of the proposed learning framework to train deep predictive policies for skilled behaviors. 
In the first experiment, a simulated PR2 robot learns how to predictively move the end-effector towards different points on a planar surface. 
The other two experiments are performed on the real PR2 robot shown in  Fig.~\ref{fig:pr2throw} and Fig.~\ref{fig:pr2grasp}. 
In these experiments, we exploit the proposed framework to train the deep policy architecture to realize two skilled behaviors; ball throwing and object grasping. 
Although the object grasping task is a well-established problem in robotics, what is referred to here is a
skilled behavior learning problem, which requires training a deep neural network policy to generate a trajectory of motor activations predictively from a single image observation snapshot.
The ball throwing task is also another instantiation of skilled behaviors which involves learning a complex sequence of motor activations for a given image observation. 
In the following, we first introduce the experimental setup (\textit{A}) and then present results from experiments on the simulated (\textit{B}) and the real robot tasks (\textit{C}).
\subsection{Experimental setup}
\label{sec:experiment_setup}
All the experiments are performed on the PR2's left arm  which is a 7-DOF manipulator. 
Each joint of the arm is controlled by a velocity PID controller which translates joint velocity commands (rad/s) to motor torques. 
The velocity commands are predictively generated by the deep predictive policy and are sent to the low-level controllers at 10 Hz. 
An external camera is mounted on top of the robot head. The camera direction is controlled by the head pan and tilt joints to view the table in front of the robot (see e.g., Fig.~\ref{fig:pr2throw}). From an RGB camera image resolution of $640\times480$ pixels, images are first cropped to $550\times350$ pixels and then down-scaled and mean-centered to $200\times200$ pixels. We use Caffe \cite{jia2014caffe} to train both the perception and behavior autoencoders and RLLAB \cite{duan2016benchmarking} with default parameters for implementations of VPG, REPS, TRPO and CEM  methods. The video and data for this work are available for download from the first author's homepage \url{http://www.csc.kth.se/~algh}. 


\subsection{Simulated reaching task}
\label{sec:simulated_tasks}
In a first experiment, a simulated PR2 robot learns to generate sequences of motor commands predictively to move its end-effector towards every point on a planar surface. Fig.~\ref{fig:simulator} demonstrates a successful trial to reach to the point indicated by the red circle. 
The network in this case consists of the policy and behavior super-layers. It receives a 2D target position on the surface as an input and generates the motor trajectory to reach that target point. 
The behavior super-layer is pre-trained as explained in Sec.~\ref{sec:method_behavior} and it is fixed during the entire experiment.  
There are two behavior models realized for the two different tasks, grasping and ball throwing. 
The reaching task utilizes the same behavior model as the grasping task. 
Both models are trained in the Gazebo simulator environment as shown in Fig.~\ref{fig:simulator}. 
The trained behavior super-layers are used for the simulation and the real robotic tasks without any further modifications.  
\begin{figure}[h]    
  \centering
    \includegraphics[width=0.47\textwidth]{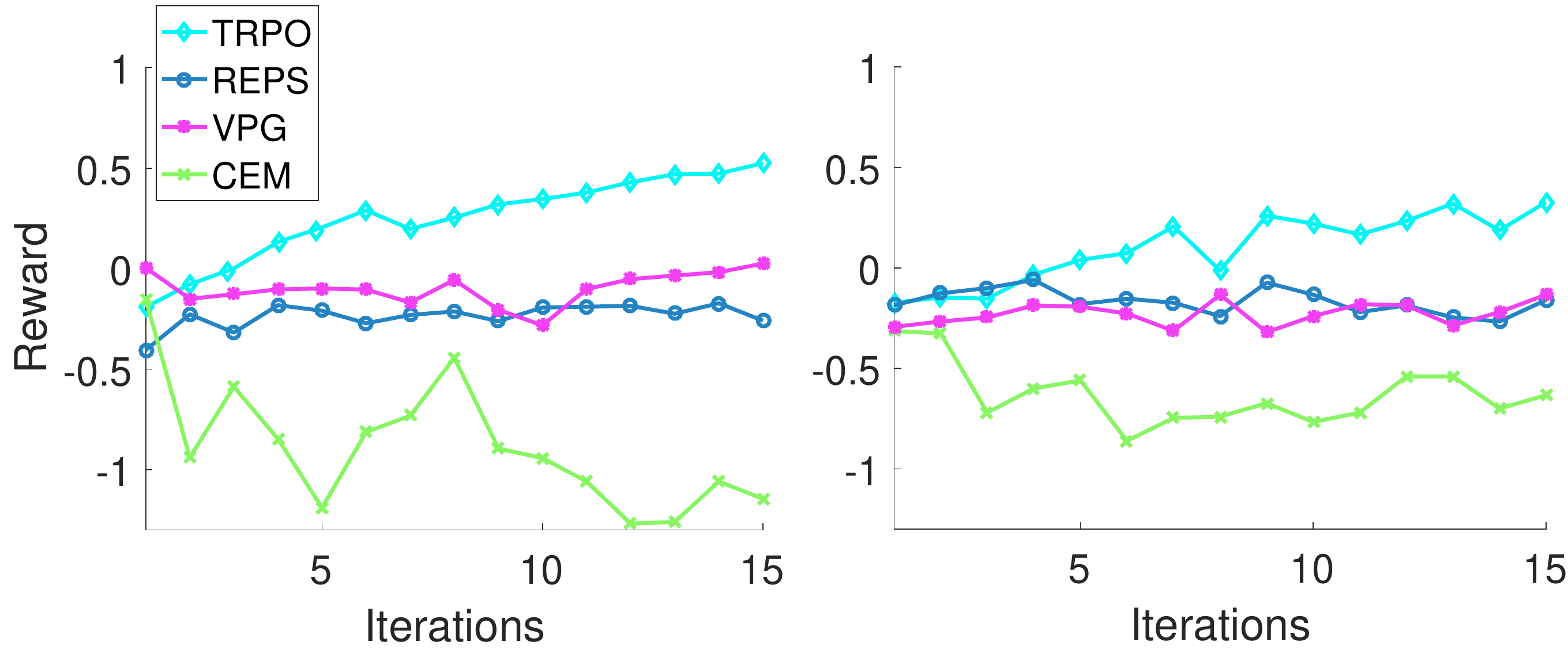}    
    \caption{The average reward for different training iterations with continuous (left) and discrete (right) reward types for the simulated reaching task.} 
    \label{fig:res_sim}
    \vspace{-0.3cm}
\end{figure}

The policy super-layer is trained with the reward signal received at the end of each episode. 
We evaluated both continuous and discrete reward values to train the policy layers. For the simulated reaching task, the continuous reward is calculated as $r =1 - ||p - p^*||^\frac{1}{2}$, where $p$ is the final 2D position of the end-effector, and $p^*$ is the target. 
The discrete reward is simply found by discretizing the continuous reward to values $+1, +0.5, -0.5$ and $-1$. 

We evaluated TRPO, REPS, CEM and VPG when used for training the policy super-layer of the reaching task. The super-layer was trained with each method for three independent trials consisting of 15 iterations.
Each iteration includes 25 reaching attempts.
Fig.~\ref{fig:res_sim} demonstrates the average reward for each iteration, for both the continuous and discrete reward types. 
It asserts that TRPO achieves the fastest learning rate compared to the rest of methods. According to the figure, VPG and REPS demonstrate similar performance, with VPG slightly better than REPS, whereas 
CEM exhibits unstable performance.
Table~\ref{table:res_sim} demonstrates the performance of the corresponding deterministic policies found after training for 15 iterations. 
The deterministic policy is found by considering the mean value of the stochastic policy.
The table evaluates the deterministic policies for target points that already exist in the training set, as well as for novel targets. 

\begin{table}[h]
\vspace{-0.2 cm}
\caption{Evaluation of the deterministic reaching policies for training and novel target points.}
\label{table:res_sim}
\renewcommand{\arraystretch}{1.3}
\centering
\begin{tabular}{|c|c|c|c|c|}
\hline
&  \multicolumn{2}{|c|}{\bfseries Training targets } & \multicolumn{2}{|c|}{\bfseries Novel targets}\\
\cline{2-5}
\cline{2-5}
& continuous & discrete & continuous & discrete\\
\hline
TRPO & 0.57 & 0.50 & 0.55 & 0.50\\
\hline
REPS & 0.01 & 0.00 & -0.01 & 0.00\\
\hline
VPG & 0.03 & 0.04 & 0.02 & 0.02\\
\hline
CEM & -1.12 & -0.24 & -1.3 & -0.26\\
\hline
\end{tabular}
\vspace{-0.2 cm}
\end{table}

As shown in this experiment, TRPO is the most suitable method to train the policy super-layer as it demonstrates the fastest learning rate. 
Furthermore, we conclude that the introduced behavior model would facilitate learning complex motor activations, requiring a reasonable number of trials with the only information provided in terms of discrete terminal rewards.
The experiment also shows that the trained action manifold renders efficient policy learning with a small sub-network possible.

\begin{figure}[h]    
  \vspace{-0.2 cm}
  \centering
    \includegraphics[width=0.28\textwidth]{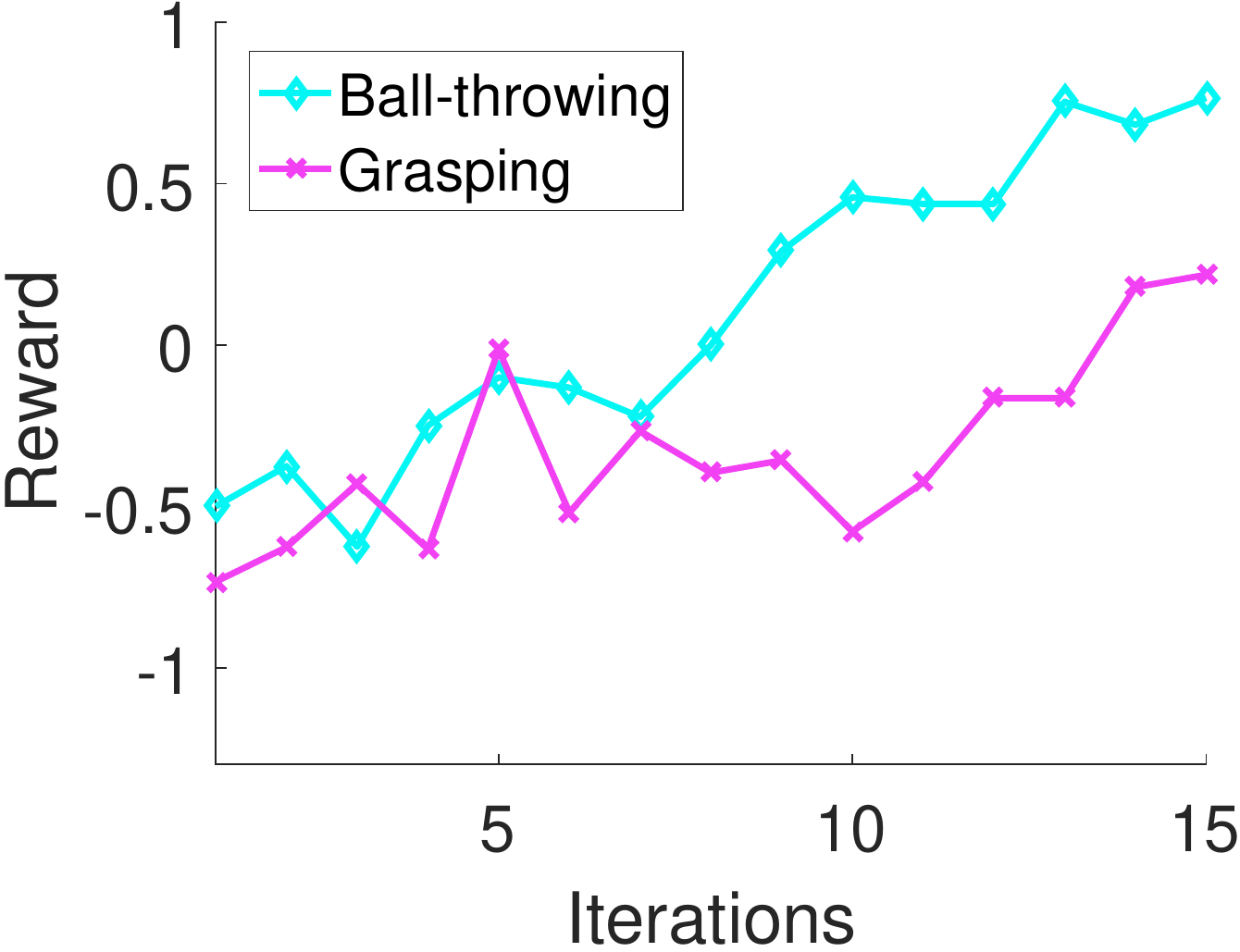}    
    \caption{Training progress for the real robot tasks} 
    \label{fig:res_real}
    \vspace{-0.3 cm}
\end{figure}
\subsection{Real robot tasks}
We trained the proposed architecture to realize the two skilled behaviors, grasping and ball throwing, on the PR2 robot.
The full network maps raw image observations to a predictive sequence of motor velocity commands through the three super-layers shown in Fig.~\ref{fig:network}. 

The perception and behavior super-layers are pre-trained as explained in Sec.~\ref{sec:method_perception} and Sec.~\ref{sec:method_behavior} and are kept fixed during the experiments. 
The perception model is shared for both tasks, while the behavior models are different. The behavior model for the grasping task is the same as the one used in the previous section for the simulated reaching task. 

As TRPO yielded the best performance in the simulations,
it is used for the training policy super-layer for the real robot tasks. 
TRPO trains the grasping policy with continuous rewards similar to the reaching task, with the difference that a successful grasp will get a reward of +2. The distance to a target is measured based on the forward kinematic model of the robot. 
For the ball throwing task, we only have access to qualitative reward values such as the ball hit the object (+2), landed closely (+1) or far (-1) to/from the object. 

Fig.~\ref{fig:res_real} illustrates the training progress for both tasks. Each task is trained for 15 iterations, with 12 action attempts each. Therefore, a total of 180 attempts are generated for each task which roughly correspond to one hour of data collection. 
The figure demonstrates the reward outcome for the stochastic policy i.e., randomly sampling the action according to the learned action distribution. 
Similarly to the previous section, we also evaluated the deterministic version of the policy and reported the results in Table~\ref{table:res_real}. 
The table demonstrates the average reward over 12 different attempts with the deterministic policies for cases when there are no visual distractors, with known distractors available during perception model training, and unknown distractors never seen before. 

\begin{table}[h]
\vspace{-0.2 cm}
\caption{Evaluation of deterministic policies for the two real robot tasks, for cases with no distractors, known and unknown distractors respectively.}
\label{table:res_real}
\renewcommand{\arraystretch}{1.3}
\centering
\begin{tabular}{|c|c|c|c|}
\hline
& \bfseries No distractors & \bfseries Known & \bfseries Unknown \\
\hline
Ball-throwing & 1.58 & 1.58 & 0.83\\
\hline
Grasping & 1.13 & 1.11 & 0.61\\
\hline
\end{tabular}
\vspace{-0.2 cm}
\end{table}

As mentioned earlier, the maximum reward is +2 which corresponds to a perfect action execution for both tasks.
Also, +1 means the task was performed nearly as well i.e., the ball landed close to the object (throwing task), or the gripper touched the object (grasping task). 
As shown in Table~\ref{table:res_real}, for the case with no distractors, both trained networks receive rewards on average greater than 1.0. For ball throwing the average reward is $1.58$ suggesting most throws hit the object. For cases with distractors seen during perception model learning, the results are close to cases with no distractors. 
However, with unknown distractors the performance is drastically reduced, suggesting that the perception model is not robust against previously unseen objects. 

\begin{figure}[h]    
  \centering
    \includegraphics[width=0.45\textwidth]{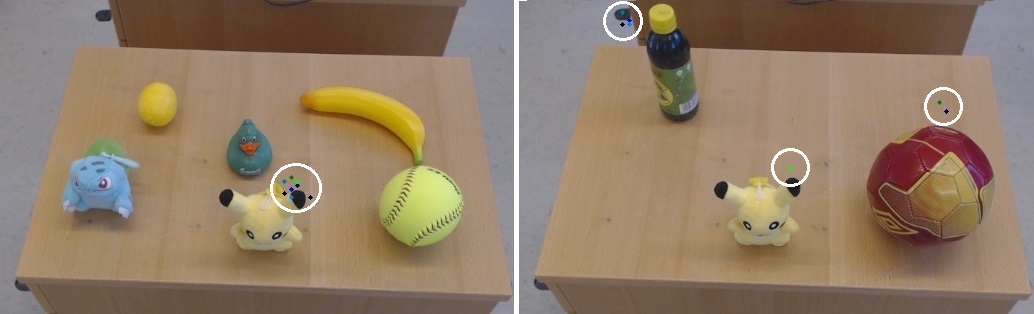}    
    \caption{Spatial feature points extracted by the perception model, for the objects observed during the training (left), and the test objects (right).} 
    \label{fig:overlaid}
    \vspace{-0.3 cm}
\end{figure}

Fig.~\ref{fig:overlaid} illustrates feature points generated by the perception model for both known and unknown distractors. As it is shown for the unknown case, some features are wrongly clustered around the distractor objects. This would result in wrong action generation and failure in the tasks. Given the limited set of training images used to train the perception model though, it is reasonable to assume that robustness can be improved with more images.

%% file: texts/conclusions.tex
\section{Conclusions and future work}
\label{sec:conclusions}
We have presented a deep predictive policy architecture with a data-efficient learning framework and evaluated it by learning skilled object grasping and ball throwing tasks on a PR2 robot. 
The architecture consists of three sub-networks referred to as perception, policy and behavior super-layers. 
Data-efficiency is achieved by training the perception and behavior super-layers with synthetic and simulated data, while only the few parameters of the policy super-layer need to be trained using real robot data. 
We experimentally demonstrated that such a neural network policy can be trained efficiently to acquire complex behaviors, such as throwing a ball and hitting a target object, with only 180 real robot training trials. 
Furthermore, we showed that the network architecture enables training of these behaviors with only qualitative terminal rewards. This is an important feature that enables a task to be trained by non-expert operators. 

We believe the proposed network architecture would gain in robustness by improving the perception model to more efficiently discard visual distractors.  
In future work, we will review different network structures with the goal of implementing a more robust perception model. 
We will also study transferability of features extracted at different layers of the model to extend the network to manipulate multiple target objects without necessarily re-training from scratch. 

Also as a part of our future studies, we intend to modify the way the behavior super-layer training samples are gathered. Currently, a blind controller generates these motor  samples. However, devising such controllers makes the system dependent on expert knowledge. To alleviate dependencies on experts, we will extend the learning framework to train the behavior super-layer with samples generated from an initial reactive controller, direct human demonstrations or motion capture systems.